# Clustering of Deep Contextualized Representations for Summarization of Biomedical Texts

**Milad Moradi, Matthias Samwald**

Institute for Artificial Intelligence and Decision Support, Center for Medical Statistics, Informatics, and Intelligent Systems, Medical University of Vienna, Vienna, Austria

`{milad.moradivastegani, matthias.samwald}@meduniwien.ac.at`

## Abstract

In recent years, summarizers that incorporate domain knowledge into the process of text summarization have outperformed generic methods, especially for summarization of biomedical texts. However, construction and maintenance of domain knowledge bases are resource-intense tasks requiring significant manual annotation. In this paper, we demonstrate that contextualized representations extracted from the pre-trained deep language model BERT, can be effectively used to measure the similarity between sentences and to quantify the informative content. The results show that our BERT-based summarizer can improve the performance of biomedical summarization. Although the summarizer does not use any sources of domain knowledge, it can capture the context of sentences more accurately than the comparison methods. The source code and data are available at https://github.com/BioTextSumm/BERT-based-Summ.

## 1 Introduction

Text summarization is the process of identifying the most important contents within a document and producing a shorter version of the text that conveys those important ideas. Many publicly available summarizers use generic features such as the position and length of sentences, the term frequency, the presence of some cue phrases, etc. to assess the importance of sentences [1]. Specifically in the biomedical domain it has been shown that these generic measures cannot be as efficient as domain-specific methods that incorporate sources of domain knowledge to represent the text on a concept-based level [2, 3]. Much effort has been invested in using sources of domain knowledge such as ontologies, taxonomies, controlled vocabularies to capture the context in which the input text appears [3-8]. These methods have improved the performance of biomedical summarization since they quantify the informative content by considering the semantics behind the sentences, rather than considering only generic features. However, building, maintaining, and utilizing sources of domain knowledge can be challenging and time-consuming tasks [9], leading the research community to develop a new generation of context-aware methods that use neural network-based language models.

In recent years, the usage of pre-trained deep language models received significant attention for a wide variety of natural language processing (NLP) tasks. In this approach, unsupervised pre-training is conducted on a large corpus of text, and the resulting model can then be 'fine-tuned' on a supervised task or can be used directly to extract numeric features for input text. The usage of deep pre-trained language models has recently obtained state-of-the-art results for a wide variety of NLP tasks [10-14].

In this paper, we propose a novel biomedical text summarizer that uses the Bidirectional Encoder Representations from Transformers (BERT) language model [14] to capture the context in which sentences appear within an input document. BERT was pre-trained on large text corpora (Wikipedia and BookCorpus) and, after a fine-tuning step, it can achieve state-of-the-art results on a wide variety of NLP tasks. It is also possible to directly extract and use contextualized embeddings learnt by BERT, without any further training or fine-tuning steps, as we do in this paper.

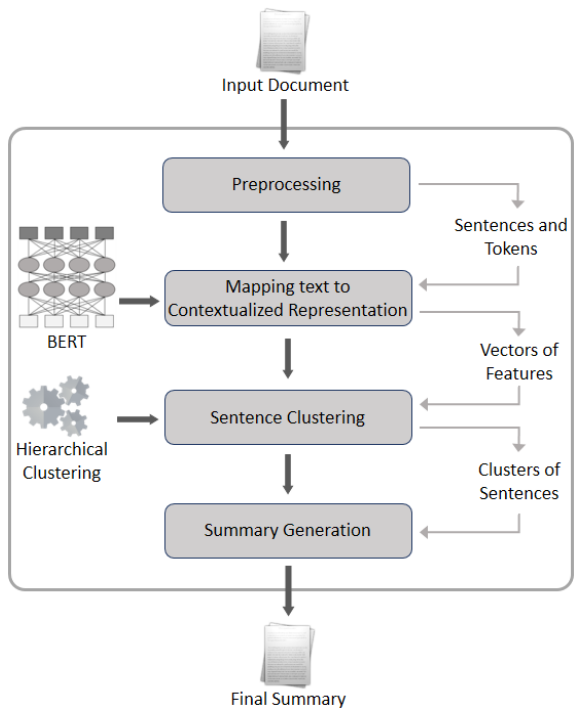

Figure 1: The overall architecture of our BERT-based biomedical text summarizer.

Utilizing the BERT language model, our summarizer computes a contextual representation, i.e. an n-dimensional vector, for every sentence. It applies a hierarchical clustering algorithm to find multiple groups of sentences, such that those sentences nearby in the vector space fall into the same cluster. The summarizer uses the contextualized embeddings to quantify the informative content of sentences and assess the similarity between them. The idea is that those sentences within the same cluster share similar context. Subsequently, the summarizer selects the most informative sentences of each cluster to generate the final summary. We evaluate the performance of our BERT-based summarizer on a corpus of biomedical scientific articles. The results show that our summarizer can improve the performance of biomedical text summarization, compared to generic methods and biomedical summarizers that utilize domain knowledge. The main contributions of this paper can be summarized as follows:

- Utilizing a pre-trained bidirectional language model for unsupervised biomedical text summarization,
- Demonstrating that the BERT-based summarizer can capture the context of sentences more accurately than the summarizers that use domain knowledge,
- Showing that clustering of deep contextualized representations can improve the performance of biomedical text summarization.

## 2 Summarization Method

Our summarization method consists of four main steps. Figure 1 illustrates the overall architecture of the summarizer.

### 2.1 Preprocessing

The summarizer performs a preprocessing step in order to prepare the input text for the subsequent steps. Those parts of the text that seem to be unimportant for appearing in the summary are discarded. These parts can vary based on the structure of the input document and user requirements. In our case, since evaluations are performed on biomedical scientific articles, the main text of input article is retained and any other parts such as headers of sections and subsections, figures and tables, etc. are discarded. This information can be added to the final summary if needed. Next, the text is split into sentences; and each sentence is split into tokens. For this purpose, we use the Natural Language Tool Kit (NLTK) library.

### 2.2 Mapping Text to Contextualized Representations

In the second step, we utilize BERT to extract contextualized embeddings. The tokens are used as the input of the feature extraction script. The output is a JSON file containing activation values of the hidden layers. Two different versions of BERT with different model sizes are currently available: BERT-Base contains 12 layers, 768 hidden units in each layer, 12 attention heads per unit, and a total number of 110 million parameters. BERT-Large contains 24 layers, 1024 hidden units in each layer, 16 attention heads per unit, and a total number of 340 million parameters. We use both BERT-Base and BERT-Large in our experiments to assess the impact of different model sizes on the quality of summaries. After the feature extraction step, each token is represented as a contextualized embedding with a size of 768 or 1024 based on the size of BERT model. Next, a contextualized representation is computed for each sentence by averaging over all the representations of tokens belonging to the sentence.

### 2.3 Sentence Clustering

The contextualized embedding of each sentence represents the context in which the sentence appears. Therefore, nearby sentences in the vector space can share similar context. The summarizer uses a clustering step to group the sentences into a number of clusters such that those in the same cluster are the most similar in terms of their representations in the vector space.

We use an agglomerative hierarchical clustering algorithm in this step. The clustering algorithm starts by specifying the number of final clusters, i.e. the parameter $K$. In each iteration, those two clusters that are the most similar (or the nearest) are merged and the number of clusters reduces by one. The similarity (or distance) between two clusters is computed by averaging over all similarity (or distance) values between each sentence of the first cluster and each sentence of the second one. The clustering algorithm proceeds until the number of clusters reaches $K$.

The similarity or distance between sentences can be computed using different measures. We run the clustering step with two widely-used measures separately, i.e. cosine similarity and Euclidean distance. Let $R=\{r_1, …, r_N\}$ and $Q=\{q_1, …, q_N\}$ be the contextualized representations of two given sentences. Cosine similarity and Euclidean distance between these two vectors are computed as follows:

$$Cosine(R, Q) = \frac{R.Q}{||R||\,||Q||} \quad (1)$$

$$Euclidean\ dist(R, Q) = \sqrt{\sum_{i=1}^{N}(r_i - q_i)^2} \quad (2)$$

At the end of this step, there is a set of clusters each one containing a set of related sentences.

### 2.4 Summary Generation

Now the summarizer needs to decide which sentences are the most relevant and informative to be included in the summary. Since those sentences within the same cluster share some important content of the input text, the summarizer selects sentences from all the clusters to cover as many important ideas as possible. Each cluster contributes to the summary in proportion to its size as follows:

$$N_i = N \frac{|C_i|}{|D|} \quad (3)$$

where $N_i$ is the number of sentences that should be selected from $i_{th}$ cluster, $N$ is the size of summary specified by the compression rate, $|C_i|$ is the size of $i_{th}$ cluster, and $|D|$ is the size of input document.

In order to select the most informative and related sentences of each cluster, a within-cluster score is computed for each sentence, as follows:

$$WCS_{i,j} = \frac{\sum_{q=1}^{|C_j|} Similarity(S_i,\ S_q)}{|C_j|} \quad (4)$$

where $WCS_{i,j}$ is the within-cluster score of $i_{th}$ sentences belonging to $j_{th}$ cluster, $|C_j|$ is the size of $j_{th}$ cluster, and $Similarity(S_i, S_q)$ is the similarity between two sentences $S_i$ and $S_q$ such that $S_i \neq S_q$. Note that the value of $Similarity(S_i, S_q)$ is computed using either measures cosine similarity or Euclidean distance, just as same as the measure used in the clustering algorithm.

Next, the summarizer ranks the sentences of each cluster based on the within-cluster scores. For each cluster $C_i$, top ranked sentences are extracted according to $N_i$. The summarizer arranges the selected sentences in the same order they appear in the input text and produces the final summary.

## 3 Experiments and Results

### 3.1 Evaluation Corpora and Metrics

We randomly retrieve 100 and 300 articles from BioMed Central to construct development and evaluation corpora, respectively. The abstract of each article is used as the model summary. This approach of creating corpora has been widely adopted in biomedical text summarization [2, 3, 5, 6]. According to [15], the size of both the corpora is large enough to allow the results to be statistically significant.

We use the ROUGE toolkit to assess the quality of summaries produced by automatic methods. Higher scores returned by ROUGE metrics refer to higher content overlap between system and model summaries. In our evaluations we use ROUGE-1 (R-1) and ROUGE-2 (R-2) metrics. R-1 and R-2 quantify the content overlap in terms of shared unigrams and bigrams, respectively.

### 3.2 Parameterization

The parameter $K$ specifies the number of final clusters in the clustering algorithm. A similarity measure is used in both the sentence clustering and summary generation steps. We assess the performance of our summarization method in different settings, varying the number of clusters in the range [2, 12] and using measures of cosine similarity and Euclidean distance separately. For brevity reasons we only report results obtained when the summarizer utilizes BERT-Large since the scores are higher than those of BERT-Base. In all experiments, the compression rate is set to 0.3.

| K | Cosine Similarity | | Euclidean distance | |
|---|---|---|---|---|
| | R-1 | R-2 | R-1 | R-2 |
| 2 | 0.7441 | 0.3329 | 0.7548 | 0.3373 |
| 3 | 0.7479 | 0.3361 | 0.7576 | 0.3397 |
| 4 | **0.7501** | **0.3394** | **0.7607** | **0.3459** |
| 5 | 0.7472 | 0.3359 | 0.7568 | 0.3402 |
| 6 | 0.7425 | 0.3317 | 0.7532 | 0.3365 |
| 7 | 0.7380 | 0.3279 | 0.7491 | 0.3330 |
| 8 | 0.7349 | 0.3251 | 0.7459 | 0.3302 |
| 9 | 0.7312 | 0.3204 | 0.7418 | 0.3283 |
| 10 | 0.7281 | 0.3184 | 0.7401 | 0.3259 |
| 11 | 0.7275 | 0.3169 | 0.7367 | 0.3217 |
| 12 | 0.7255 | 0.3136 | 0.7324 | 0.3202 |

Table 1: ROUGE scores obtained by the BERT-based summarizer in parameterization experiments.

| | R-1 | R-2 |
|---|---|---|
| BERT-based summarizer | **0.7639** | **0.3481** |
| CIBS | 0.7501 | 0.3345 |
| Bayesian summarizer | 0.7453 | 0.3301 |
| SUMMA | 0.7264 | 0.3179 |
| TexLexAn | 0.7111 | 0.3107 |

Table 2: ROUGE scores obtained by our BERT-based summarizer and the comparison methods.

Table 1 presents the ROUGE scores obtained by the summarizer using different settings. The scores are presented for both the cosine similarity and Euclidean distance. The summarizer obtains the highest scores when $K=4$. For smaller values of $K$, some important sentences are merged with sentences of larger clusters; they may lose their chance for inclusion in the summary. In this case, some informative sentences may be excluded from summaries, leading to a decrease in the quality of summarization. On the other hand, when higher values are assigned to $K$, some unimportant sentences leave large clusters, construct a new cluster, and contribute to the summary. In this case, a number of non-informative sentences may appear in the summary, decreasing the scores.

### 3.3 Comparison to other Summarizers

We evaluate the performance of our summarization method against four comparison methods, i.e. CIBS [2], the Bayesian biomedical summarizer [3], SUMMA [16], and TexLexAn[1]. CIBS uses UMLS concepts in combination with itemset mining and clustering to identify and extract important sentences. The Bayesian summarizer applies a probabilistic heuristic on concepts to produce an informative summary. SUMMA and TexLexAn employ generic features such as the length and position of sentences, the frequency of terms, the presence of cue terms, etc. Table 2 presents ROUGE scores obtained by the methods. The BERT-based summarizer reports the highest scores. Compared to the scores obtained by the comparison methods, the BERT-based summarizer can significantly ($p<0.05$) improve the performance of biomedical text summarization according to a Wilcoxon signed-rank text with a confidence interval of 95%.

## 4 Conclusion

The results show that contextualized embeddings learnt by BERT can be effectively used for biomedical text summarization. It is shown that this type of contextual representations can convey the context of sentences more accurately than the comparison methods that utilize sources of domain knowledge. This study can be an initial step toward employing this type of language models for developing domain-specific NLP systems, especially in biomedical text summarization. To extend our research we plan to utilize this type of language models trained on biomedical text corpora, and investigate their usefulness in biomedical text summarization. Future work may include the usage of contextual representations to address problems such as biomedical named entity recognition, question answering, and information extraction that need to accurately capture the context of text.

## References

[1] M. Gambhir and V. Gupta, "Recent automatic text summarization techniques: a survey," *Artificial Intelligence Review,* vol. 47, pp. 1-66, 2016.

---
[1] http://texlexan.sourceforge.net/

[2] M. Moradi, "CIBS: A biomedical text summarizer using topic-based sentence clustering," *Journal of Biomedical Informatics,* vol. 88, pp. 53-61, 2018/12/01/ 2018.

[3] M. Moradi and N. Ghadiri, "Different approaches for identifying important concepts in probabilistic biomedical text summarization," *Artificial Intelligence in Medicine,* vol. 84, pp. 101-116, 2018.

[4] R. Mishra, J. Bian, M. Fiszman, C. R. Weir, S. Jonnalagadda, J. Mostafa*, et al.*, "Text summarization in the biomedical domain: a systematic review of recent research," *Journal of biomedical informatics,* vol. 52, pp. 457-467, 2014.

[5] L. Plaza, A. Díaz, and P. Gervás, "A semantic graph-based approach to biomedical summarisation," *Artificial intelligence in medicine,* vol. 53, pp. 1-14, 2011.

[6] M. Moradi and N. Ghadiri, "Quantifying the informativeness for biomedical literature summarization: An itemset mining method," *Computer Methods and Programs in Biomedicine,* vol. 146, pp. 77-89, 2017.

[7] M. Moradi, "Frequent Itemsets as Meaningful Events in Graphs for Summarizing Biomedical Texts," in *2018 8th International Conference on Computer and Knowledge Engineering (ICCKE)*, 2018, pp. 135-140.

[8] M. Moradi, "Concept-based single- and multi-document biomedical text summarization," Isfahan University of Technology, 2017.

[9] W. W. Fleuren and W. Alkema, "Application of text mining in the biomedical domain," *Methods,* vol. 74, pp. 97-106, 2015.

[10] J. Turian, L. Ratinov, and Y. Bengio, "Word representations: a simple and general method for semi-supervised learning," presented at the Proceedings of the 48th Annual Meeting of the Association for Computational Linguistics, Uppsala, Sweden, 2010.

[11] M. E. Peters, W. Ammar, C. Bhagavatula, and R. Power, "Semi-supervised sequence tagging with bidirectional language models," *arXiv preprint arXiv:1705.00108,* 2017.

[12] A. Radford, K. Narasimhan, T. Salimans, and I. Sutskever, "Improving language understanding by generative pre-training," *URL https://s3-us-west-2. amazonaws. com/openai-assets/research-covers/languageunsupervised/language understanding paper. pdf,* 2018.

[13] M. E. Peters, M. Neumann, M. Iyyer, M. Gardner, C. Clark, K. Lee*, et al.*, "Deep contextualized word representations," *arXiv preprint arXiv:1802.05365,* 2018.

[14] J. Devlin, M.-W. Chang, K. Lee, and K. Toutanova, "Bert: Pre-training of deep bidirectional transformers for language understanding," *arXiv preprint arXiv:1810.04805,* 2018.

[15] C.-Y. Lin, "Looking for a few good metrics: Automatic summarization evaluation-how many samples are enough?," in *NTCIR*, 2004.

[16] H. Saggion, "SUMMA: A robust and adaptable summarization tool," *Traitement Automatique des Langues,* vol. 49, 2008.